\documentclass[conference, a4paper]{IEEEtran}
\IEEEoverridecommandlockouts

\usepackage{amsmath,amssymb,amsfonts}
\usepackage{algorithmic}
\usepackage{graphicx}
\usepackage{textcomp}
\usepackage{times}
\usepackage{epsfig}
\PassOptionsToPackage{hyphens}{url}
\usepackage{adjustbox}
\usepackage{balance}
\usepackage[ruled,vlined]{algorithm2e}
\usepackage[caption=false]{subfig}

\usepackage{multicol}
\usepackage{multirow}

\usepackage{xurl}
\usepackage[pagebackref=true,breaklinks=true,colorlinks,bookmarks=false]{hyperref}

\def\BibTeX{{\rm B\kern-.05em{\sc i\kern-.025em b}\kern-.08em
    T\kern-.1667em\lower.7ex\hbox{E}\kern-.125emX}}

\begin{document}

\title{Fighting Deepfakes by Detecting GAN DCT Anomalies}

\author{
\IEEEauthorblockN{Oliver Giudice}
\IEEEauthorblockA{\textit{Department of Mathematics}\\ \textit{and Computer Science} \\
\textit{University of Catania}\\
Catania, Italy \\
giudice@dmi.unict.it} \\
\and
\IEEEauthorblockN{Luca Guarnera}
\IEEEauthorblockA{\textit{Department of Mathematics}\\ \textit{and Computer Science} \\
\textit{University of Catania}\\
Catania, Italy \\
luca.guarnera@unict.it} 
\and
\IEEEauthorblockN{Sebastiano Battiato}
\IEEEauthorblockA{\textit{Department of Mathematics}\\ \textit{and Computer Science} \\
\textit{University of Catania}\\
Catania, Italy \\
battiato@dmi.unict.it}
}

\maketitle
\begin{abstract}
To properly contrast the Deepfake phenomenon the need to design new Deepfake detection algorithms arises; the misuse of this formidable A.I. technology brings serious consequences in the private life of every involved person. {State-of-the-art} proliferates with solutions using deep neural networks to detect a fake multimedia content but unfortunately these algorithms appear to be neither generalizable nor explainable.
However, traces left by Generative Adversarial Network (GAN) engines during the creation of the Deepfakes can be detected by analyzing ad-hoc frequencies. 
For this reason, in this paper we propose a new pipeline able to detect the so-called GAN Specific Frequencies (GSF) representing a unique fingerprint of the different generative architectures.
By employing Discrete Cosine Transform (DCT), anomalous frequencies were detected. The $\beta$ statistics inferred by the AC coefficients distribution have been the key to recognize GAN-engine generated data. Robustness tests were also carried out in order to demonstrate the effectiveness of the technique using different attacks on images such as JPEG Compression, mirroring, rotation, scaling, addition of random sized rectangles. Experiments demonstrated that the method is innovative, exceeds the {state of the art} and also give many insights in terms of explainability.
\end{abstract}

\begin{IEEEkeywords}
deepfake detection; Generative Adversarial Networks; multimedia forensics; image forensics
\end{IEEEkeywords}

\section{Introduction}
 Artificial Intelligence (AI) techniques to generate synthetic media and their circulation on the network led to the birth, in 2017, of the Deepfake phenomenon: altered (or created) multimedia content by ad-hoc machine learning generative models, e.g., the Generative Adversarial Network (GAN)~\cite{goodfellow2014generative}. Images and videos of famous people, available on different media like TV and Web, could appear authentic at first glance, but they may be the result of an AI process which delivers very realistic results. In this context the 96\% of these media are porn (deep porn)~\cite{vaccari2020deepfakes}. If we think that anyone could be the subject of this alteration we can understand how a fast and reliable solution is needed to contrast the Deepfake phenomenon. Most of the techniques already proposed in literature act as a “black box” by tuning ad-hoc deep architectures to distinguish “real” from “fake” images generated by specific GAN machines. It seems not easy to find a robust detection method capable of working in the wild; even current solutions need a considerable amount of computing power. 
Let’s assume that any generative process based on GAN, presents an automated operating principle, resulting from a learning process. In~\cite{guarnera2020deepfake}, it has been already demonstrated that it is possible to attack and retrieve the signature on the network’s de-convolutional layers; in this paper a method to identify any anomaly of the generated “fake” signal, only partially highlighted in some preliminary studies~\cite{guarnera2020preliminary, zhang2019detecting} is presented.  The Fourier domain demonstrated to be prone and robust into understanding semantic at superordinate level~\cite{oliva2001modeling}. Spatial domain has been recently further investigated by~\cite{xu2020learning,xu2019training,NEURIPS2019_b05b57f6} to gain robustness and exploit related biasness~\cite{rahaman2019spectral}. To improve the efficiency, the $8\times8$ DCT has been exploited, by employing similar data analysis made in \cite{farinella2015representing, ravi2016semantic} and extracting simple statistics of the underlying distribution~\cite{lam2000mathematical}. The final classification engine based on gradient boosting, properly manages and isolates the GAN Specific Frequencies ($GSF$),
 of each specific architecture, a sort of fingerprint/pattern, outperforming state-of- the-art methods. In this paper a new “white box” method of Deepfake detection called CTF-DCT (Capture the Fake by DCT Fingerprint Extraction) is proposed, based on the analysis of the Discrete Cosine Transform (DCT) coefficients. Experiments on Deepfake images of human faces proved that a proper signature of the generative process is embedded on the given spatial frequencies. In particular we stress the evidence, that such kind of images, have in common global shape and main structural elements allowing to isolate artefacts that are not only unperceivable but also capable to discriminate between the different GANs. Finally, the robust classifier is able to demonstrate its generalizing ability in the wild even on Deepfakes not generated by GAN-engines demonstrating the ability to catch artefacts related to reenactment forgeries.

\begin{figure*}[t!]
    \centering
    \includegraphics[width=15cm]{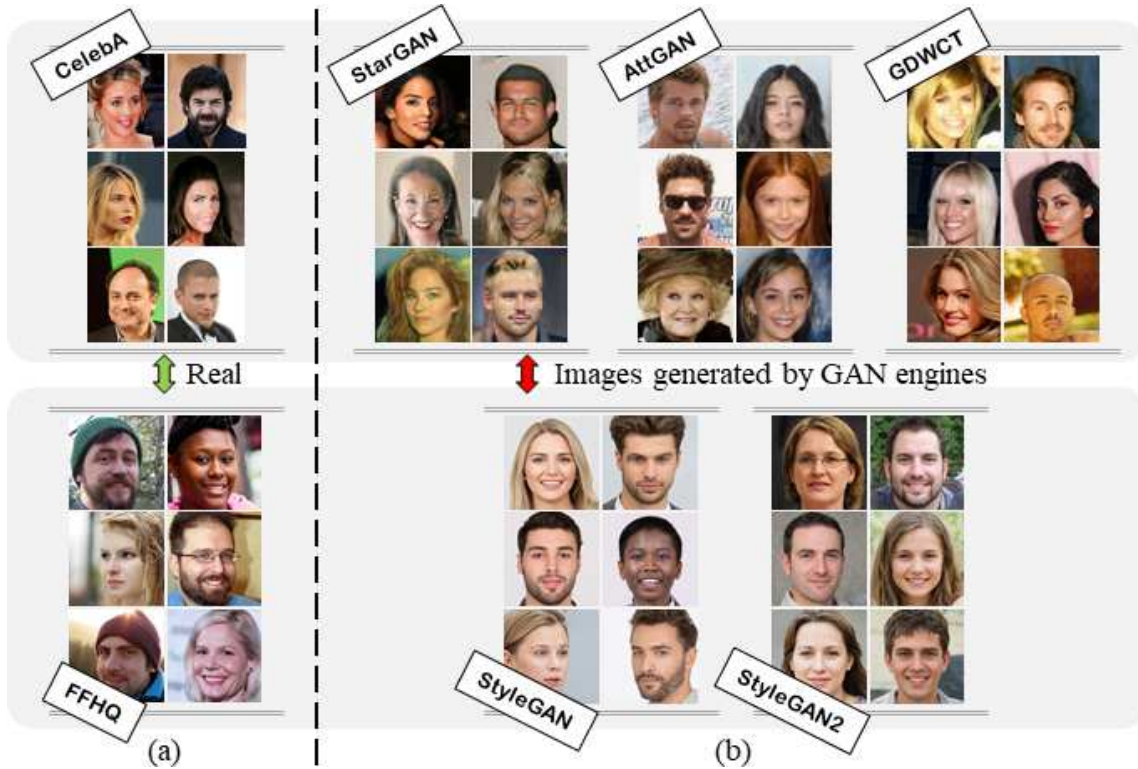}
    \caption{Example of real (\textbf{a}) and deepfake datasets (\textbf{b}) used in our experiments. The CelebA dataset was used to generate human face images with the StarGAN, AttGAN and GDWCT architectures. The FFHQ dataset was used to generate human face images with the StyleGAN and StyleGAN2 architectures.}

   \label{fig:datasets}
\end{figure*}

The main contributions of this research are the following: 
\begin{itemize}
    \item A new high-performance Deepfake face detection method based on the analysis of the AC coefficients calculated through the Discrete Cosine Transform, which delivered not only great generalization results but also impressive classification results with respect to previous published works. The method does not require computation via GPU and “hours” of training to perform Real Vs Deepfake classifications;
    \item The detection method is “explainable” (white-box method). Through a simple estimation of the characterizing parameters of the Laplacian distribution, we are able to detect those anomalous frequencies generated by various Deepfake architectures;
    \item Finally, the detection strategy was attacked to simulated situations in the wild. Mirroring, scaling, rotation, addition of random size rectangles, position and color were applied to the images, also demonstrating the robustness of the proposed method and the ability to perform well even on video dataset never taken into account during training.
\end{itemize}

The paper is organized as follows: Section~\ref{sec:related} presents the state-of-the-art of Deepfake generation and detection methods. The proposed approach is described in Section \ref{sec:method}. 
Section~\ref{sec:discussionGSF}, a discussion of GSF is reported. Experimental results, robustness test and comparison with competing methods are reported in Section~\ref{sec:experimentalResults}. Section~\ref{sec:conclusion} concludes the paper with suggestions for future works.

\section{Related Works}

\label{sec:related}

AI-synthetic media are generally created by techniques based on GANs, firstly introduced by Goodfellow et al.~\cite{goodfellow2014generative}. GANs train two models simultaneously: a generative model $G$, that captures the data distribution, and a discriminative model $D$, able to estimate the probability that a sample comes from the training data rather than from $G$. The training procedure for $G$ is to maximize the probability of $D$ making a mistake thus resulting to a min-max two-player game. 

An overview on Media forensics with particular focus on Deepfakes has been recently proposed in~\cite{tolosana2020deepfakes, verdoliva2020media}.

Five of the most famous and effective architectures in state-of-the-art for Deepfakes facial images synthesis were taken into account (StarGAN~\cite{choi2018stargan}, StyleGAN~\cite{karras2019style}, StyleGAN2~\cite{karras2020analyzing}, ATTGAN~\cite{he2019attgan} GDWCT~\cite{cho2019image}) used in our experiments as detailed below.

\subsection{Deepfake Generation Techniques of Faces}

StarGAN \cite{choi2018stargan}, proposed by Choi et al., is a method capable of performing image-to-image translations on multiple domains (such as change hair color, change gender, etc.) using a single model. Trained on two different types of face datasets---CELEBA~\cite{liu2015deep} containing 40 labels related to facial attributes such as hair color, gender and age, and RaFD dataset~\cite{langner2010presentation} containing 8 labels corresponding to different types of facial expressions (``happy'', ``sad'', etc.)---this architecture, given a random label as input (such as hair color, facial expression, etc.), is able to perform an image-to-image translation operation with impressive visual result. 

An interesting study was proposed by He et al. \cite{he2019attgan} with a framework called AttGAN in which an attribute classification constraint is applied in the latent representation to the generated image, in order to guarantee only the correct modifications of the desired attributes.

Another style transfer approach is the work of Cho et al.~\cite{cho2019image}, proposing a group-wise deep whitening-and coloring method (GDWCT) for a better styling capacity. They used CELEBA, Artworks~\cite{zhu2017unpaired}, cat2dog~\cite{lee2018diverse}, Ink pen and watercolor classes from Behance Artistic Media (BAM)~\cite{wilber2017bam}, and Yosemite datasets~\cite{zhu2017unpaired} as datasets improving not only computational efficiency but also quality of generated images.

Finally, one of the most recent and powerful methods regarding the  entire-face synthesis is the Style Generative Adversarial Network architecture or commonly called \mbox{StyleGAN \cite{karras2019style}}, where, by means of mapping points in latent space to an intermediate latent space, the framework controls the style output at each point of the generation process. Thus, StyleGAN is capable not only of generating impressively photorealistic and high-quality photos of faces, but also offers control parameters in terms of the overall style of the generated image at different levels of detail. While being able to create realistic pseudo-portraits, small details might reveal the fakeness of generated images.  To correct those imperfections, Karras et al. made some improvements to the generator (including re-designed normalization, multi-resolution, and regularization methods) proposing StyleGAN2~\cite{karras2020analyzing} obtaining extremely realistic faces. 
Figure~\ref{fig:datasets} shows an example of facial images created by five different generative architectures.

\subsection{Deepfake Detection Techniques}

Almost all currently available strategies and methods for Deepfake detection are focused on anomalies detection trying to find artefact and traces of the underlying generative process.
The Deepfake images could contain a pattern pointed out by the analysis of anomalous peaks appearing in the spectrum in the Fourier domain. Zhang et al.~\cite{zhang2019detecting} analyze the artefacts induced by the up-sampler of GAN pipelines in the frequency domain. The authors proposed to emulate the synthesises of GAN artefacts. Results obtained by the spectrum-based classifier greatly improves the generalization ability, achieving very good performances in terms of binary classification between authentic and fake images. Also Durall et al.~\cite{durall2019unmasking} presented a method for Deepfakes detection based on the analysis in the frequency domain. The authors combined high-resolution authentic face images from different public datasets (CELEBA-HQ data set~\cite{karras2017progressive}, Flickr-Faces-HQ data set~\cite{karras2019style}) with fakes ($100$K Faces project ({\url{https://generated.photos/}}, accessed on 14/02/2021), this person does not exist~({\url{https://thispersondoesnotexist.com/}}, accessed on 14/02/2021)), creating a new dataset called Faces-HQ. By means of naive classifiers they obtained effective results in terms of overall accuracy of detection.

Wang et al.~\cite{wang2019fakespotter} proposed FakeSpotter, a new method based on monitoring single neuron behaviors to detect faces generate by Deepfake technologies. The authors used in the experiments CELEBA~\cite{liu2015deep} and FFHQ~(\url{https://github.com/NVlabs/ffhq-dataset}, accessed on 14/02/2021) images (real datasets of faces) and compared Fakespotter with \mbox{Zhang et al.~\cite{zhang2019detecting}} obtaining an average detection accuracy of more than 90\% on the four types of fake faces: Entire Synthesis~\cite{karras2017progressive, karras2020analyzing}, Attribute Editing~\cite{choi2018stargan, liu2019stgan}, Expression Manipulation~\cite{karras2019style, liu2019stgan}, DeepFake~\cite{rossler2019faceforensics++, li2020celeb}.

The work proposed by Jain et al.~\cite{jain2020detecting} consists of a framework called DAD-HCNN which is able to distinguish unaltered images from those that have been retouched or generated through different GANs by applying a hierarchical approach formed by three distinct levels. 
The last level is able to identify the specific GAN model (STARGAN~\cite{choi2018stargan}, SRGAN~\cite{ledig2017photo}, DCGAN~\cite{radford2015unsupervised}, as well as the Context Encoder~\cite{pathak2016context}). 
Liu et al.~\cite{liu2020global} proposed an architecture called Gram-Net, where, through the analysis of a global image texture representations, they managed to create a robust fake image detection. The results of the experiments, done both with Deepfake (DCGAN, StarGAN, PGGAN, StyleGAN) and real images (CelebA, CelebA-HQ, FFHQ), demonstrate that this new type of detector delivers effective results.

Recently, a study conducted by Hulzebosch~\cite{hulzebosch2020detecting} describes that the CNN  solutions presented till today for Deepfake detection are limited to lack of robustness, generalization capability and explainability, because they are extremely specific to the context  in  which  they  were  trained  and,  being  very  deep, tend to extract the underlying semantics from images. For this reason, in literature new algorithms capable to find the Deepfake content without the use of deep architectures were proposed. As described by \mbox{Guarnera et al.~\cite{guarnera2020deepfake, guarnera2020fighting}}, the current GAN architectures leaves a pattern (through convolution layers) that characterizes that specific neural architecture. In order to capture this forensic trace, the authors used the Expectation-Maximization Algorithm~\cite{moon1996expectation} obtaining features able to distinguish real images from Deepfake ones. Without the use of deep neural networks, the authors exceeded state-of-the-art in terms of accuracy in the real Vs Deepfake classification test, using not only Deepfake images generated by common GAN architectures, but also testing images generated by modern FaceApp mobile application.

Differently from the described approaches, in this paper, the possibility to capture the underlying pattern of a possible Deepfake is investigated  extracting the discriminative features through the DCT transform.

\section{The CTF Approach}
\label{sec:method}

\begin{figure*}[t!]
    
    \includegraphics[width=\linewidth]{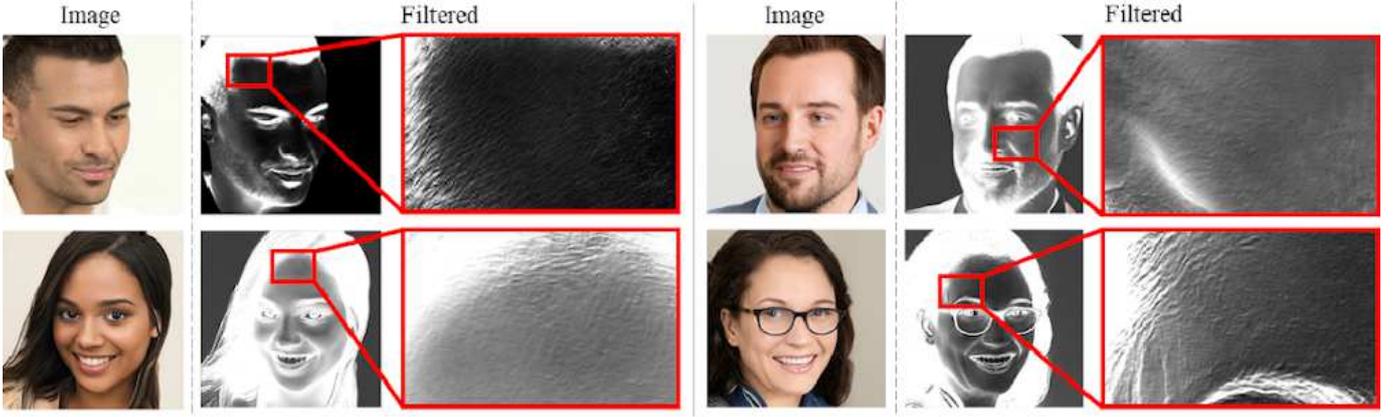}
    \caption{Example of image generated by StyleGAN properly filtered to highlight patterns resulting from the generative process.}
    \label{fig:StyleGAN}
\end{figure*}
\begin{figure*}[t!]
    
    \includegraphics[width=\linewidth]{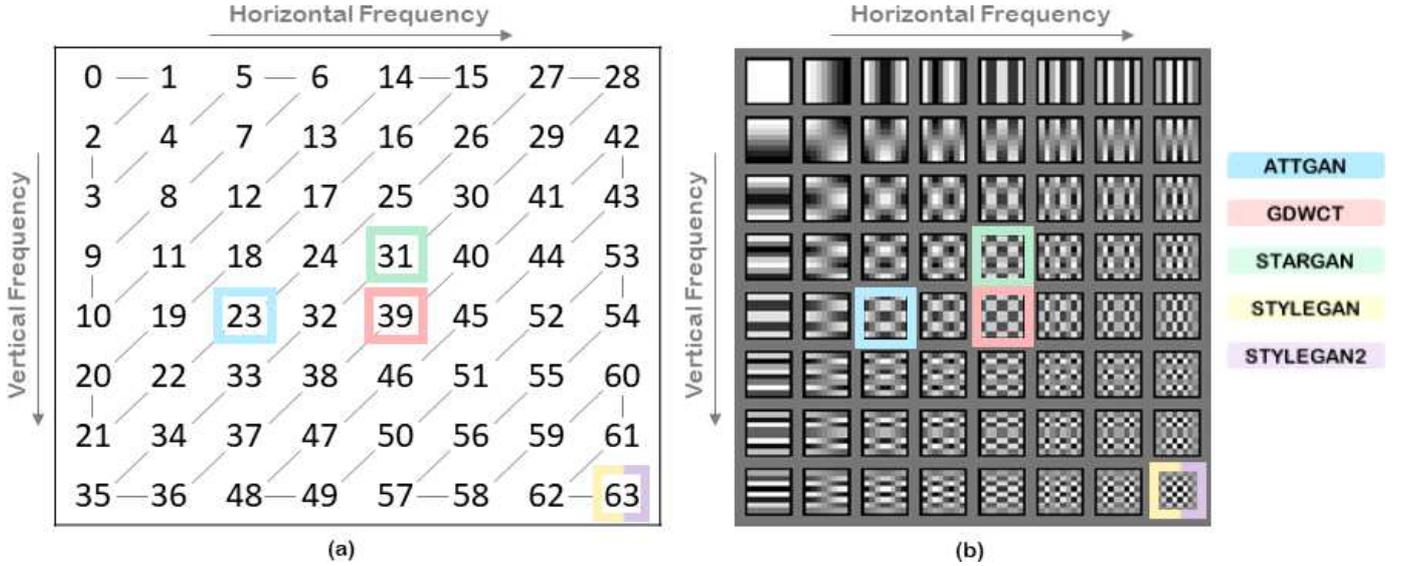}
    \caption{GSF that identify the generative architectures. (\textbf{a}) Zig-zag order after DCT transform. \mbox{(\textbf{b}) DCT} $8\times8$ frequencies.}
    \label{fig:zigzag}
\end{figure*}
 
In \cite{hulzebosch2020detecting}, Dutch law enforcement experts were tasked with discriminating between images from the FFHQ dataset and StyleGAN images,  which were created   starting from FFHQ. The results reached only the 63\% of accuracy while state-of the-art \mbox{methods \cite{guarnera2020fighting, wang2019fakespotter}} are able to deliver a better outcome. Algorithms were used for extracting black-box features that likely are not related to the visible domain but are somehow encoding anomalies strictly dependent on the way Deepfakes are generated. In particular, a refined evaluation of the StyleGAN images, shows that some abnormal patterns are visible in the most structured part of the images (e.g., skin, hair, etc.). Given such a repetitive pattern, which would have to be subsisting on the middle bands of the Deepfake image frequency spectrum,  a frequency-based approach might be able to detect it and describe it.  To this end, the CTF approach will take place by leveraging more than a decade of JPEG compression pipeline studies employing DCT block-based processing, which is effectively used for many computer vision and image forensics tasks not strictly related to compression \mbox{itself~\cite{jing2004face,ravi2016semantic, farinella2015representing,thai2015camera,lam2004analysis}.}

The CTF approach transform and analyse images on the DCT domain in order to detect the most discriminant information related to the pattern shown in Figure \ref{fig:StyleGAN} which is typical of the employed Generative Model (e.g., GAN).

Let $I$ be a digital image. Following the JPEG pipeline, $I$ is divided into non-overlapping blocks of size $8 \times 8$. The Discrete Cosine Transform (DCT) is then applied to each block, formally:
\begin{equation}
	\label{eq:DCT1}
	F[u,v] =  {\frac{1}{4}}C(u)C(v)\Biggl[\sum\limits_{x=0}^7\sum\limits_{y=0}^7I[x,y]\mbox{cos}{({a})}\mbox{cos}{({b})}\Biggl]
\end{equation}
where
$a = \frac{(2x+1)u\pi}{16}$, $b = \frac{(2y+1)v\pi}{16}$,
$C(u) = \bigg \{
\begin{array}{rl}
\frac{1}{\sqrt{2}} & u = 0 \\
1 & u > 0\\
\end{array} $ and $C(v) = \bigg \{
\begin{array}{rl}
\frac{1}{\sqrt{2}} & v = 0 \\
1 & v > 0\\
\end{array} $.

For each $8 \times 8$ block, the 64 elements $F[u,v]$ form the DCT coefficients. They are sorted into a zig-zag order starting from the top-left element to the bottom right (Figure~\ref{fig:zigzag}). The DCT coefficient at position 0 is called DC and represents the average value of pixels in the block. All others coefficients namely AC, corresponds to specific bands of frequencies.

Given all the DCT transformed $8 \times 8$ blocks of $I$, it is possible to assess some statistics for each DCT coefficient. By applying evidence reported in~\cite{lam2000mathematical}, the DC coefficient can be modelled with a Gaussian distribution while the AC coefficients were demonstrated to follow a zero-centred Laplacian distribution described by:
\begin{equation}
	\label{eq:laplacian}
	\begin{split}
	P(x) =  \frac{1}{2\beta}exp\Biggl(\frac{-|x-\mu|}{\beta} \Biggl)
	\end{split}
\end{equation}
with $\mu=0$ and $\beta = \sigma/\sqrt{2}$ is the scale parameter where $\sigma$ corresponds to the standard deviation of the AC coefficient distributions.
The proposed approach is partially inspired by~\cite{farinella2015representing} where a GMM (Gaussian Mixture Model) over different $\beta$ values has been properly adopted for scene classification at superordinate level.

\begin{figure}[t!]
    \centering
    \includegraphics[width=9cm]{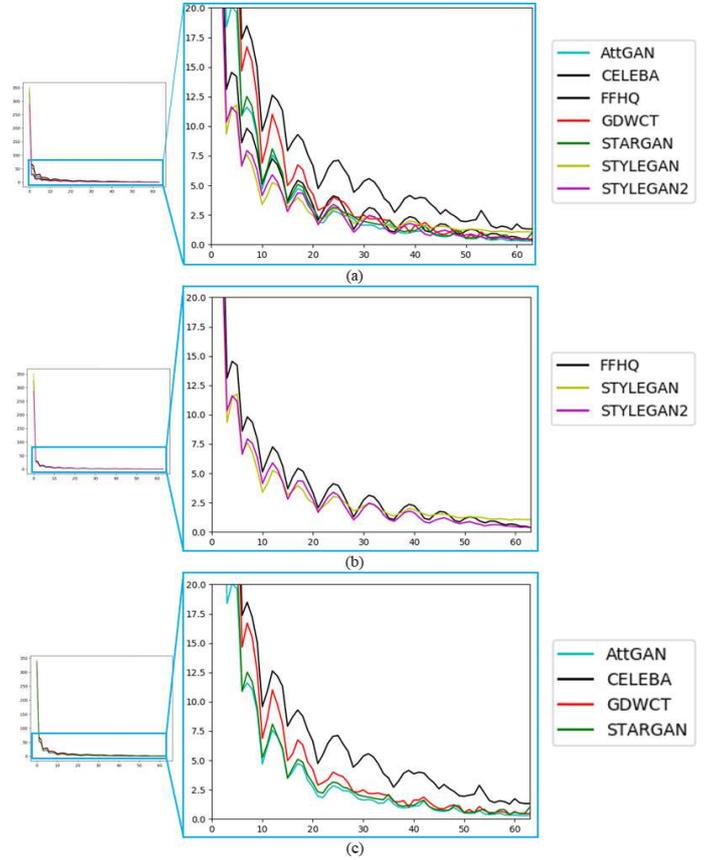}
    \caption{Plot of $\beta$ statistics of each involved dataset. The average $\beta$ value for each $i$-th coefficient is reported. (\textbf{a}) Shows the average $\beta$ trend of all datasets (real and deepfake); (\textbf{b}) Shows the average $\beta$ trend of StyleGAN and StyleGAN2 compared to the real image dataset used for their creation (FFHQ); (\textbf{c}) Shows the average $\beta$ trend of StarGAN, AttGAN and GDWCT compared to the real image dataset used for their creation (CelebA). For each plot, the abscissa axis represents the 64 coefficients of the $8\times8$ block, while the ordinate axis are the respective inferred $\beta$ values (in our case the average of the $\beta$ values computed for all images of the respective datasets).}
    \label{fig:betas}
\end{figure}

An accurate estimation of such $\beta$ values for each coefficient and  involved GAN-engine, is crucial for the purpose achievement.
Figure~\ref{fig:betas}  graphically summarizes the statistical trend of the $\beta$-values of each involved datasets showing empirically the intrinsic discriminative power devoted to distinguish almost univocally images generated by GAN-engines or picked-up from real datasets.
Let $\vec \beta_I = \{\beta_{I_1}, \beta_{I_2}, \dots, \beta_{I_N}\} $ with $N = 63$ (DC coefficient is excluded) the corresponding feature vector of the image $I$. We exploited related statistics on different image-datasets $DT_g$ with \textit{g = \{StarGAN, AttGAN, GDWCT, StyleGAN, StyleGAN2, CelebA, FFHQ\}}.

For the sake of comparisons in our scenario we evaluated pristine images generated by StarGAN~\cite{choi2018stargan}, AttGAN~\cite{he2019attgan}, GDWCT~\cite{cho2019image}, StyleGAN~\cite{karras2019style}, StyleGAN2~\cite{karras2020analyzing}, and genuine images extracted by CelebA~\cite{liu2015deep} and FFHQ. E.g., $DT_{StyleGAN}$ represents all the available images generated by StyleGAN engine.
For each image-set $DT_g$ let's consider the following representation:
\begin{equation}
\label{eq:betamatrix}
    \beta_{DT_{g}} = 
    \begin{pmatrix}
  \vec\beta_{1} \\
  \vec\beta_{2} \\
  \vdots  \\
    \vec\beta_{|DT_g|}
 \end{pmatrix}
 =
 \begin{pmatrix}
  \beta_{1,1} & \beta_{1,2} & \cdots & \beta_{1,63} \\
  \beta_{2,1} & \beta_{2,2} & \cdots & \beta_{2,63} \\
  \vdots  & \vdots  & \ddots & \vdots  \\
  \beta_{|DT_g|,1} & \beta_{|DT_g|,2} & \cdots & \beta_{|DT_g|,63}
 \end{pmatrix}
\end{equation}
where $|.|$ is the number of images in $DT_{g}$. 
For sake of simplicity, in the forthcoming notation all dataset $DT_g$ have been selected to have the same size $|DT_g| = K$. Note that $\beta_{DT_{g}}$ have been normalised w.r.t. each column.
To extract GSF we first computed the distance among the involved AC distributions modelled by $\beta_{DT_g}$ for each dataset.
We computed a $\chi^2$ distance as follows:
\begin{equation}
\label{eq:chi}
    \begin{split}
    \chi^2(\beta_{DT_i}, \beta_{DT_j}) = 
    \sum_{r=1}^{K} \frac{(\beta_{DT_i}[r,c] - \beta_{DT_j}[r,c])^2}{\beta_{DT_j}[r,c]} \quad
    \\
    \textrm{with}\quad c=1,\dots,63
    \end{split}
\end{equation}
where $i,j \in g, i \neq j$, 
$c$ is the column which corresponds to the AC coefficient and $r$ are the rows in (\ref{eq:betamatrix}) that represents all $\vec \beta_I$ features. The distance $\chi^2(\beta_{DT_i}, \beta_{DT_j})$ is a vector with size {of} $63$.
Finally, it is possible to define the GAN Specific Frequency ($GSF$) as follows:

\begin{equation}
\label{eq:gsf}
    GSF_{DT_{i},DT_{j}} = \underset{c}{\text{ argmax }}  \sum_{r=1}^{K} \frac{\left(\beta_{D T_{i}}[r, c]-\beta_{D T_{j}}[r, c]\right)^{2}}{\beta_{D T_{j}}[r, c]}
\end{equation}
where, $i,j \in g, i \neq j$. GSF allow to realize a one-to-one evaluation between image sets. 

\begin{figure*}[t!]
    
    \includegraphics[width=\linewidth]{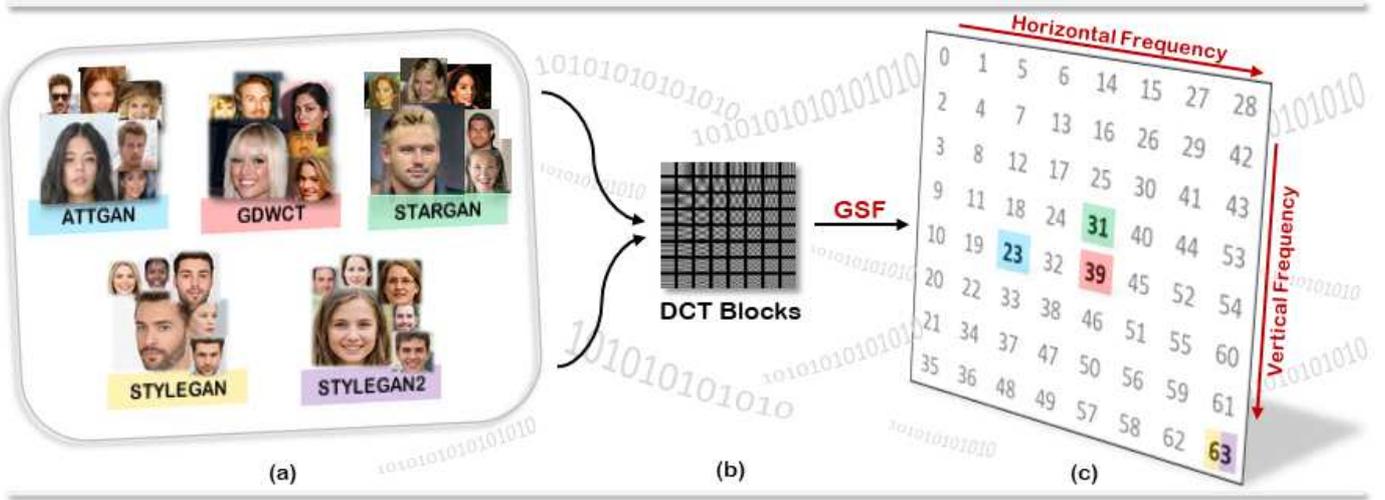}
    \caption{CTF-DCT approach: (\textbf{a}) Dataset used for our experiments; (\textbf{b}) Discrete Cosine Transform (DCT) of a given image at each 8x8 blocks; (\textbf{c}) GAN Specific Frequencies (GSF) that identify involved GAN architectures.}
    \label{fig:genericalGANc}
\end{figure*}

Practically, the most discriminative DCT frequency is selected among two datasets in a greedy fashion and, as proven by experiments, there is no need to add further computational steps (e.g., frequency ranking/sorting, etc.). In Figure~\ref{fig:genericalGANc}c, GSF computed for a set of pair of image-sets, are highlighted just to provide a first toy example where \mbox{$200$ images} ($K = 200$) for each set have been employed. Specifically AttGAN, StarGAN and GDWCT were compared with the originating real image-set (CelebA) and for the same reason StyleGAN and StyleGAN2 were compared with FFHQ.

The $\beta$ values as described in the experiments, are very discriminative when it comes to deepfake detection. Figure~\ref{fig:betas} shows the average trend of $\beta$ of all images from the respective Real and Deepfake datasets. It is interesting to analyze the trend of $\beta$ of the Deepfake images compared to the statistics of the Real dataset used for the generation task. \mbox{Figure~\ref{fig:betas}c} shows StarGAN, AttGAN, and GDWCT Vs CelebA. All DCT coefficients are sorted in terms of JPEG zigzag order as shown in Figure~\ref{fig:zigzag}a.
It is worth noting that if we consider even only one of the $\beta$ values we can roughly establish if an image is a deepfake simply by properly thresholding specific frequencies according to the definition of GSF (\mbox{Equation~(\ref{eq:gsf})}). Please note that the discriminative power of the GSFs, even if in some sense they bring energies due to the involved DCT frequencies as demonstrated by the detection results, are not fully dependent by the involved resolution.

\section{Datasets Details}
\label{sec:datasedetails}

Two datasets of real face images were used for the employed experimental phase: CelebA and FFHQ. Different Deepfake images were generated considering StarGAN, GDWCT, AttGAN, StyleGAN and StyleGAN2 architectures. In particular, CelebA images were manipulated using pre-trained models available on Github, taking into account StarGAN, GDWCT and AttGAN. Images of StyleGAN and StyleGAN2 created through FFHQ were downloaded ad detaled in the following:
\begin{itemize}
    \item CelebA (CelebFaces Attributes Dataset): a large-scale face attributes dataset with more than 200 K celebrity images, containing 40 labels related to facial attributes such as hair color, gender and age. The images in this dataset cover large pose variations and background clutter. The dataset is composed by $178\times218$ JPEG images.
    \item FFHQ (Flickr-Faces-HQ): is a high-quality image dataset of human faces with variations in terms of age, ethnicity and image background. The images were crawled from Flickr and automatically aligned and cropped using dlib~\cite{dlib09}. The dataset is composed by high-quality $1024\times1024$ PNG images.
    \item StarGAN is able to perform Image-to-image translations on multiple domains using a single model. Using CelebA as real images dataset, every image was manipulated by means of a pre-trained model~(\url{https://github.com/yunjey/stargan}, accessed on 14/02/2021) obtaining a final resolution equal to $256\times256$.
    \item GDWCT is able to improve the styling capability. Using CelebA as real images dataset, every image was manipulated by means of a pre-trained model~(\url{https://github.com/WonwoongCho/GDWCT}, accessed on 14/02/2021) obtaining a final resolution equal to $216\times216$.
    \item AttGAN is able to transfers facial attributes with constraints. Using CelebA as real images dataset, every image was manipulated by means of a pre-trained model~(\url{https://github.com/LynnHo/AttGAN-Tensorflow}, accessed on 14/02/2021) obtaining a final resolution equal to $256\times256$.
    \item StyleGAN is able to transfers semantic content from a source domain to a target domain characterized by a different style. Images have been generated considering FFHQ as dataset in input with $1024\times1024$ resolution (\url{https://github.com/NVlabs/stylegan}, accessed on 14/02/2021).
    \item StyleGAN2 improves STYLEGAN quality with the same task. Images have been generated considering FFHQ as dataset in input with $1024\times1024$ resolution~(\url{https://github.com/NVlabs/stylegan2}, accessed on 14/02/2021).
\end{itemize}

For all the carried out experiments, 3000 Deepfake images for each GAN architecture and 3000 from CelebA and FFHQ were collected and divided into training and test set as will be reported in experimental dedicated Sections. Figure~\ref{fig:datasets} shows some examples of the employed real and Deepfake images.

\section{Discussion on GSF}
\label{sec:discussionGSF}

\begin{figure*}[t!]
    \includegraphics[width=\linewidth]{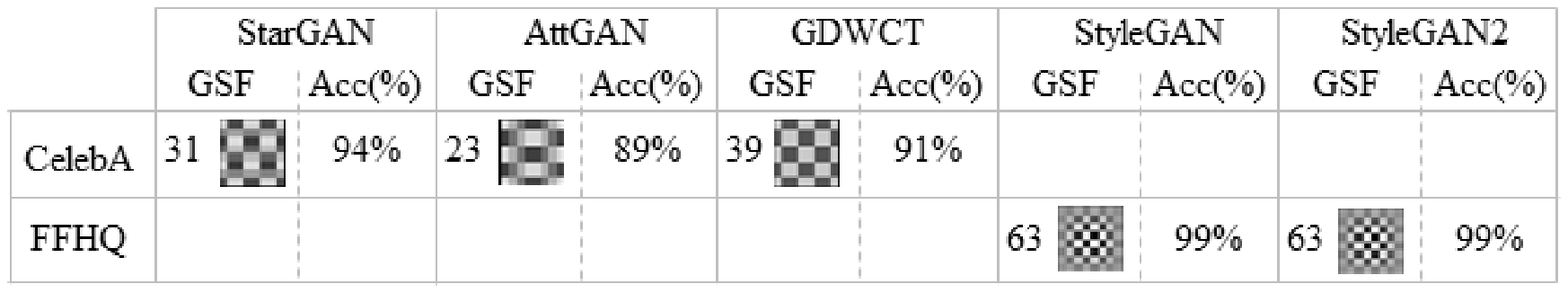}
    \caption{Average Accuracy results (\%) obtained for the binary classification task employing only the $GSF$. 700 images were employed for testing, 200 images for training, 5-fold cross validated, classes are balanced.}
    \label{fig:GSFirstResults}
\end{figure*}

Although differentiating between a Deepfake and a real image could be easy, given the high accuracy values demonstrated by state-of-the-art methods~\cite{wang2020cnn}, it becomes difficult when the test is carried out on fake images obtained from a specific set of real images: for instance differentiating between FFHQ images and StyleGAN ones, which are generated from FFHQ images, is more difficult than differentiating StyleGAN vs. CelebA images. As a matter of fact, state-of-the-art methods like Fakespotter \cite{wang2019fakespotter} employs for training, mixed sets of Deepfake and real images. Results are then unbalanced by the extremely-easy-to-spot-difference like CelebA vs. StyleGAN. This can be demonstrated by means of $GSF$ analysis.

Through $GSF$ it is possible to perform a one-to-one test between sets of images. This was carried out specifically for the harder case as described before: taking 200 images for each set, $GSF$ was calculated for each pair of image sets, whose values obtained are shown in Figure \ref{fig:zigzag}b. In particular, AttGAN, StarGAN and GDWCT were compared with the starting real images (CelebA) and for the same reason StyleGAN and StyleGAN2 were compared with FFHQ.

Torralba et al.~\cite{zhou2017scene} demonstrated that scenes semantic-visual components are captured precisely with analogous statistics on spectral domain used also to build fast classifiers of scenes~\cite{farinella2015representing}. In this sense, the comparison between images that represent close-ups of faces showing the some overall visual structure raising extremely similar statistical characteristics of AC coefficients and their $\beta$ values. This allows the $GSF$ analysis to focus on the unnatural anomalies introduced by the convolutional generative process typical of Deepfakes. To demonstrate the discriminative power of the $GSF$ a simple binary classifier (logistic regression) was trained using the $\beta$ (e.g., that corresponds the set of values of a given column/coefficient in Equation~(\ref{eq:betamatrix})) of the corresponding $GSF$ as unique feature. 

For all the experiments carried out, the number of collected images has been equally set considering $K = 3000$. In particular the classifier was trained using only the 10\% of the entire dataset, while the remaining part was used as test set.
For each binary classification test, the simple classification solution obtained the results shown in Figure~\ref{fig:GSFirstResults}. 
Results demonstrated that Deepfakes are easily detectable by just looking at the $\beta$ value of the $GSF$ for that specific binary test. 
This is empirically found to be discriminative (wider range of values) than expected on natural images, given the semantic context of facial images. This finding is what state-of-the-art is exploiting with much more complex and computational intensive solutions. For instance, Fakespotter~\cite{wang2019fakespotter}, at a first step compares real against fake images and finds these unnatural frequencies with an ad-hoc trained CNN. As a matter of fact, frequencies found are different for forgeries made with Photoshop which certainly do not bear traces of convolution and for this reason they are easily discriminated from the Deepfake images.

\begin{figure*}[t!]
    
    \includegraphics[width=\linewidth]{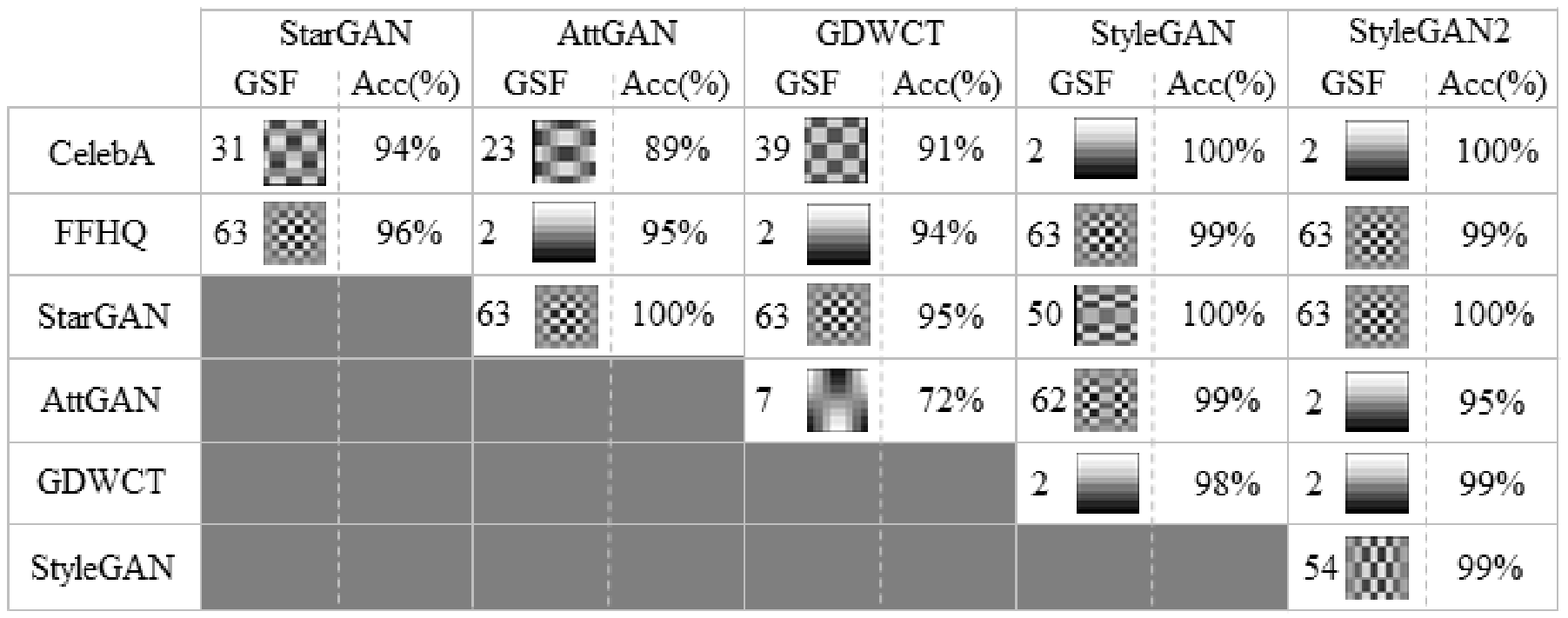}
    \caption{GSF and classification accuracy results (\%) obtained for each binary classification task.}
    \label{fig:GsfAll}
\end{figure*}

\begin{figure}[t!]
{\captionsetup{position=bottom,justification=centering}
     \subfloat[\label{subfig:AttGAN}]{%
       \includegraphics[width=0.3\linewidth]{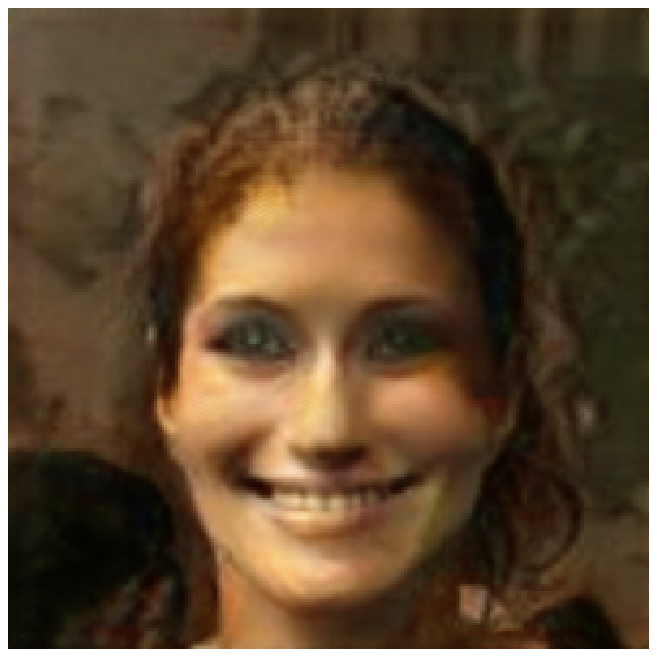}
     }
     \hfill
     \subfloat[\label{subfig:Fourier1}]{%
       \includegraphics[width=0.3\linewidth]{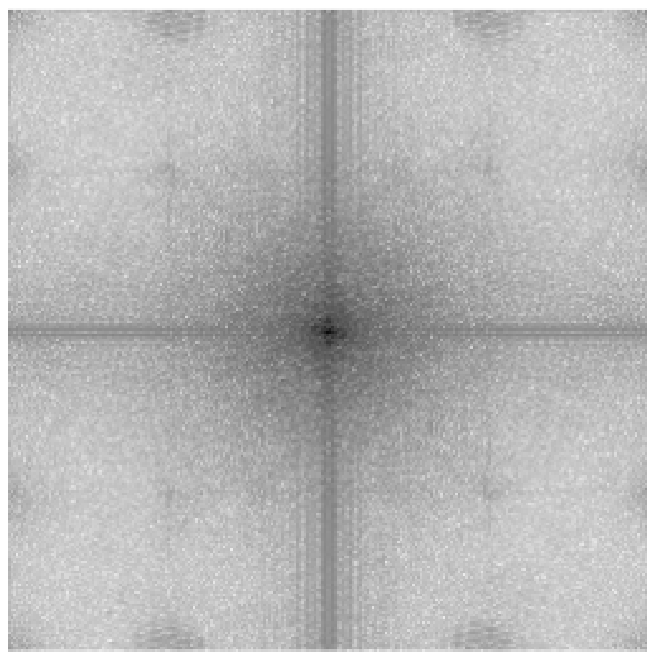}
     }
    \hfill
     \subfloat[\label{subfig:Fourier2}]{%
       \includegraphics[width=0.3\linewidth]{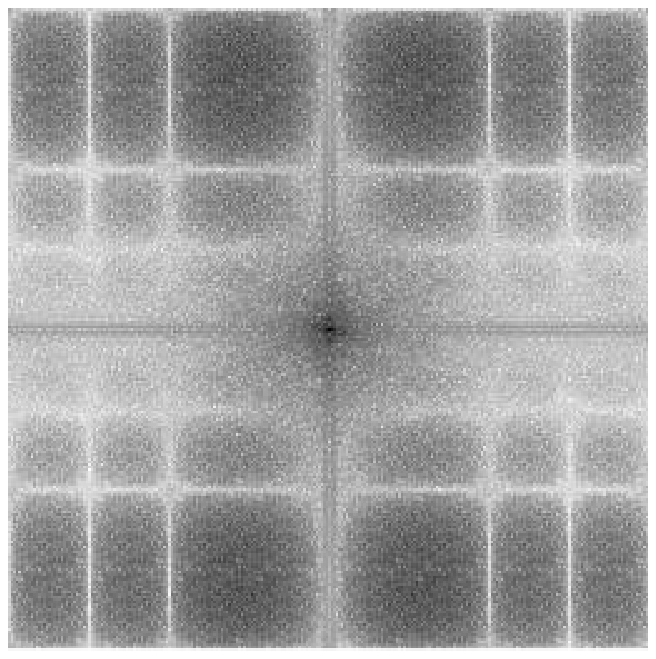}
     }}
    \hfill
     \caption{Abnormal frequencies inspection. (\textbf{a}) Image example from the StarGAN dataset; (\textbf{b}) Fourier Spectra of the input image (\textbf{a}); (\textbf{c}) Abnormal frequency shown by means of $GSF$ amplification. Examples showing $GSF$ amplification of the involved generation processes will be extensively reported on supplementary material.}
     \label{fig:Hough-PnP}
\end{figure}

As already stated, the combination of different resolution and frequency bands image-sets is the major problem  encountered in the state of the art methods, while the most problematic issue is  differentiating the original images from the transformed Deepfake. Let's take into account FFHQ vs. STYLEGAN: a task in which even the human being had difficulties~\cite{hulzebosch2020detecting}. 
Applying GSF analysis among all involved proper datasets, we obtain impressive generalization results as reported in Figure~\ref{fig:GsfAll}.
Further demonstration of the importance of the $GSF$ will be visual. In addition to the anomalies visually identified in Figure \ref{fig:Hough-PnP}, in Guarnera et al. \cite{guarnera2020preliminary} the authors already identified some strange components in the Fourier spectrum. Given an image from a specific image-set, after having computed the $GSF$ (Figure \ref{fig:genericalGANc}), it is possible for sake of explainability, to apply the following amplification process: to multiply in the DCT domain each DCT coefficient different from the $GSF$ by a value $k_1$ (with $0<k_1\leq1$) while the coefficient corresponding to the $GSF$ by a value $k_2$ (with $k_2>1$). Figure~\ref{fig:Hough-PnP} shows an example of such amplification procedure with $k_1={0.1}$ and $k_2 = 100$. This operation will create an image where the $GSF$ is amplified. Figure~\ref{fig:Hough-PnP} shows that the original Fourier Spectrum and the amplified one share the same abnormal frequency appearance. Thus, $GSF$ becomes an explanation of those anomalies with a clear boost of forensics  analysis.

It has to be noted that the $GSF$ approach described in this section is a great instrument to white-box GAN-generated image processing. A $GSF$ is able to identify a set of {GAN-generated images}. On the other hand, it is not enough to properly being employed in the wild or against fakes not generated by neural approaches. For this reason, in the following section, we ``finalize'' the approach by presenting a more robust and complete feature vector but, on the other hand, we will lose explainability.

\subsection{Finalizing the CTF Approach}

Given the ability of the $GSF$ to make one-to-one comparisons even between image-sets of GANs it is possible to use it to resolve further discrimination issues. Figure \ref{fig:genericalGANc} shows that the two StyleGANs actually have the same $GSF_{StyleGAN,FFHQ} = GSF_{StyleGAN2,FFHQ} = 63$, while $GSF_{StyleGAN,StyleGAN2} = 54$ was obtained (Figure~\ref{fig:GsfAll}). Also upon this $GSF$ it is possible to train a classifier that quickly obtains an accuracy value in the binary test between StyleGAN and StyleGAN2 close to 99\%.

The $GSF$ analysis can be exploited to give explainability to unusual artifacts and behaviors that appear in the Fourier domain of Deepfakes. Obviously, using only the corresponding $\beta$ to $GSF$ can be reductive for a scenario in the wild and this is the reason why the CTF approach will be completed by means of a robust classifier which will be outlined in the next section. Instead of using only the corresponding $\beta$ to the $GSF$, it will employ a feature vector with all 63 $\beta$, consequently  used as input to a Gradient Boosting classifier \cite{bishop2006pattern} and tested in a noisy context that includes a number of plausible attacks on the images under analysis.  
Gradient Boosting was selected as the best classifier for data and the following hyper-parameters were selected by means of a $10\%$ of the dataset employed as validation set. We selected the following hyper parameters: \emph{number-of-estimators} = 100, \emph{learning-rate} = 0.6, $max_{depth} = 2$.

The robust classifier thus created, fairly identify the most probable GAN from which the image has been generated, providing hints for “visual explainability”. By considering the growing availability of Deepfakes to attack people reputation such aspects become fundamental to assess and validate forensics evidence.
All the employed data and code will be publicly available after the review process at a public link.

\section{Experimental Results}
\label{sec:experimentalResults}

\begin{figure*}[t!]
    
    \includegraphics[width=\linewidth]{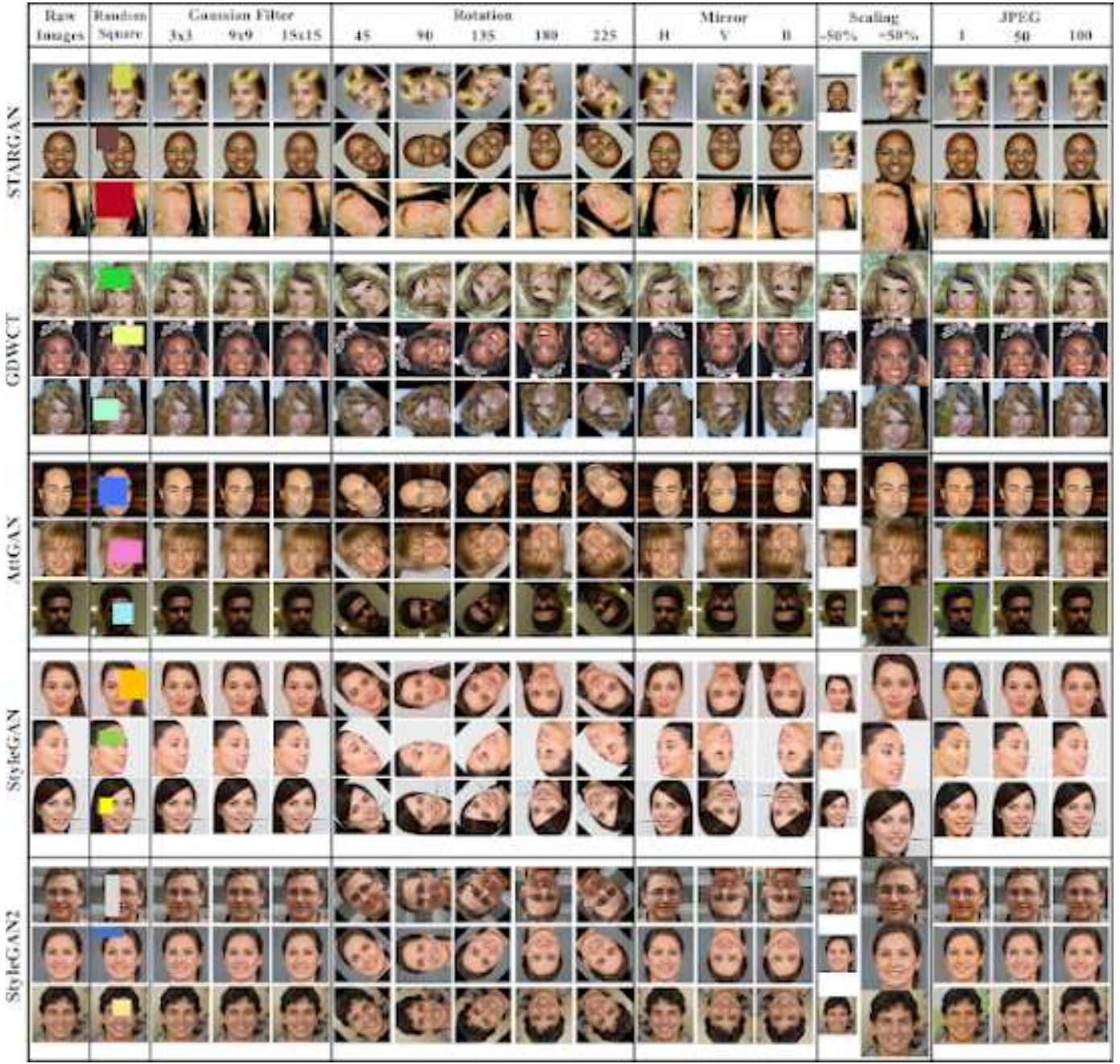}
    \caption{Examples of ATTGAN, GDWCT, STARGAN, STYLEGAN, STYLEGAN2 images in which we applied different attacks: Random Square, Gaussian Blur, Rotation, Mirror, Scaling and JPEG Compression. They were also applied in the real dataset (CelebA and FFHQ).}
    \label{fig:attacks}
\end{figure*}

In this section experimental results are presented. Primarily, to finalize the CTF approach, a robust classifier was trained and tested by means of several attacks on images and consequently tested in a different scenario, namely the FaceForensics++ dataset of Deepfake videos \cite{rossler2019faceforensics++}. The above-mentioned deepfake dataset is used only during the testing phase to classify real Vs deepfake.
3000 real and fake images were collected to train the ``robust classifier'' for the validation, employing only the  10\% of the entire dataset while the remaining part was used as test set.
Multiple attack types augmented the dataset; Figure~\ref{fig:attacks} provides examples of images after each attack. Cross-validation was carried out.

\subsection{Testing with Noise}
\label{sec:robustness}

All the images collected in the corresponding $DT$ have been put through different kinds of attacks as addition of a random size rectangle, position and color, Gaussian blur, rotation and mirroring, scaling and various JPEG Quality Factor compression (QF), in order to demonstrate the robustness of the CTF approach.

As shown in Table~\ref{tab:noise_tests} this type of attacks do not destroy the $GSF$ obtaining high accuracy values.

\begin{table*}[t!]
\caption{Percentage of Precision, Recall, F1-score and accuracy obtained in the robustness test. “Raw Images” shows the results without the attack process. For the “Real” column the CelebA and FFHQ datasets were considered. Different attacks were carried out in the datasets: Random square; Gaussian filter with different kernel size ($3x3,9x9,15x15$); Rotations with $degree=\{45,90,135,180,225\}$; Mirror with Horizontal (H), Vertical (V) and Both (B) ways; Scaling ($+50\%, -50\%$); JPEG Compression with different Quality Factor ($QF=\{1,50,100\}$).}
\centering
  \begin{adjustbox}{max width=\linewidth}
\begin{tabular}{|cc|c|c|c|c|c|c|c|c|c|ccc|c|c|c|c|c|c|c|}
\hline
\multicolumn{1}{|l}{}                                                      & \multicolumn{1}{l|}{} & \multicolumn{3}{c|}{\textbf{Real}} & \multicolumn{3}{c|}{\textbf{AttGAN}} & \multicolumn{3}{c|}{\textbf{GDWCT}} & \multicolumn{3}{c}{\textbf{StarGAN}}                                & \multicolumn{3}{c|}{\textbf{StyleGAN}} & \multicolumn{3}{c|}{\textbf{StyleGAN2}} & \textbf{Overall}  \\ \cline{3-20}
\multicolumn{1}{|l}{}                                                      & \multicolumn{1}{l|}{} & \textbf{Prec}     & \textbf{Rec}    & \textbf{F1}& \textbf{Prec}     & \textbf{Rec}    & \textbf{F1}      & \textbf{Prec}     & \textbf{Rec}    & \textbf{F1}    & \multicolumn{1}{c|}{\textbf{Prec}} & \multicolumn{1}{c|}{\textbf{Rec}} & \textbf{F1}  & \textbf{Prec}     & \textbf{Rec}    & \textbf{F1}      & \textbf{Prec}     & \textbf{Rec}    & \textbf{F1}       & \textbf{Accuracy} \\ \hline
\multicolumn{2}{|c|}{\textbf{Raw Images}}                                                                   & 99       & 97     & 98    & 99       & 100     & 99     & 98       & 98      & 98    & \multicolumn{1}{c|}{99}   & \multicolumn{1}{c|}{100} & 100 & 99        & 98       & 99     & 98        & 100      & 99      & 99       \\ \hline
\multicolumn{2}{|c|}{\textbf{Random Square}}                                                                & 98       & 94     & 96    & 90       & 96      & 93     & 92       & 89      & 91    & \multicolumn{1}{c|}{100}  & \multicolumn{1}{c|}{98}  & 99  & 98        & 99       & 98     & 99        & 99       & 99      & 96       \\ \hline
\multirow{3}{*}{\begin{tabular}[c]{@{}c@{}}\textbf{Gaussian}\\\textbf{Filter}\end{tabular}} & \textbf{3x3}                   & 98       & 95     & 96    & 83       & 88      & 86     & 89       & 92      & 91    & \multicolumn{1}{c|}{92}   & \multicolumn{1}{c|}{86}  & 89  & 97        & 98       & 98     & 99        & 99       & 99      & 93       \\ \cline{3-21} 
                                                                           & \textbf{9x9}                   & 98       & 99     & 98    & 62       & 59      & 60     & 70       & 79      & 74    & \multicolumn{1}{c|}{59}   & \multicolumn{1}{c|}{53}  & 56  & 99        & 98       & 99     & 98        & 99       & 98      & 81       \\ \cline{3-21} 
                                                                           & \textbf{15x15}                 & 100      & 97     & 98    & 58       & 64      & 61     & 72       & 64      & 68    & \multicolumn{1}{c|}{55}   & \multicolumn{1}{c|}{53}  & 54  & 98        & 99       & 98     & 95        & 100      & 97      & 80       \\ \hline
\multirow{5}{*}{\textbf{Rotation}}                                                  & \textbf{45$^\circ$}                    & 97       & 93     & 95    & 85       & 82      & 83     & 92       & 98      & 95    & \multicolumn{1}{c|}{84}   & \multicolumn{1}{c|}{84}  & 84  & 97        & 99       & 98     & 99        & 98       & 98      & 92       \\ \cline{3-21} 
                                                                           & \textbf{90$^\circ$}                    & 98       & 99     & 98    & 95       & 99      & 97     & 98       & 93      & 95    & \multicolumn{1}{c|}{100}  & \multicolumn{1}{c|}{99}  & 99  & 99        & 98       & 98     & 99        & 99       & 99      & 98       \\ \cline{3-21} 
                                                                           & \textbf{135$^\circ$}                   & 95       & 96     & 96    & 85       & 83      & 84     & 97       & 94      & 96    & \multicolumn{1}{c|}{83}   & \multicolumn{1}{c|}{86}  & 85  & 96        & 95       & 96     & 96        & 97       & 97      & 92       \\ \cline{3-21} 
                                                                           & \textbf{180$^\circ$}                   & 98       & 94     & 96    & 95       & 100     & 97     & 97       & 95      & 96    & \multicolumn{1}{c|}{99}   & \multicolumn{1}{c|}{100} & 99  & 98        & 99       & 99     & 100       & 99       & 99      & 98       \\ \cline{3-21} 
                                                                           & \textbf{225$^\circ$}                   & 96       & 95     & 95    & 88       & 85      & 87     & 96       & 96      & 96    & \multicolumn{1}{c|}{86}   & \multicolumn{1}{c|}{89}  & 88  & 96        & 97       & 97     & 97        & 97       & 97      & 93       \\ \hline
\multirow{3}{*}{\textbf{Mirror}}                                                    & \textbf{H}                    & 99       & 96     & 98    & 99       & 100     & 99     & 98       & 99      & 98    & \multicolumn{1}{c|}{99}   & \multicolumn{1}{c|}{100} & 99  & 99        & 99       & 99     & 100       & 100      & 100     & 99       \\ \cline{3-21} 
                                                                           & \textbf{V}                     & 99       & 96     & 98    & 99       & 100     & 99     & 97       & 99      & 98    & \multicolumn{1}{c|}{99}   & \multicolumn{1}{c|}{100} & 100 & 99        & 99       & 99     & 100       & 100      & 100     & 99       \\ \cline{3-21} 
                                                                           & \textbf{B}                     & 99       & 94     & 97    & 98       & 100     & 99     & 97       & 99      & 98    & \multicolumn{1}{c|}{99}   & \multicolumn{1}{c|}{100} & 100 & 99        & 99       & 99     & 100       & 100      & 100     & 99       \\ \hline
\multirow{2}{*}{\textbf{Scaling}}                                                   & \textbf{+50\%}                 & 99       & 98     & 99    & 94       & 95      & 95     & 95       & 93      & 94    & \multicolumn{1}{c|}{98}   & \multicolumn{1}{c|}{99}  & 99  & 99        & 99       & 99     & 99        & 100      & 99      & 97       \\ \cline{3-21} 
                                                                           & \textbf{-50\%}                 & 74       & 95     & 84    & 77       & 66      & 71     & 74       & 72      & 73    & \multicolumn{1}{c|}{81}   & \multicolumn{1}{c|}{77}  & 79  & 82        & 85       & 84     & 90        & 81       & 85      & 80       \\ \hline
\multirow{3}{*}{\textbf{JPEG}}                                                      & \textbf{1}                     & 78       & 69     & 73    & 63       & 65      & 64     & 59       & 67      & 63    & \multicolumn{1}{c|}{59}   & \multicolumn{1}{c|}{57}  & 58  & 78        & 83       & 80     & 84        & 80       & 82      & 70       \\ \cline{3-21} 
                                                                           & \textbf{50}                    & 93       & 95     & 94    & 98       & 99      & 98     & 87       & 80      & 83    & \multicolumn{1}{c|}{84}   & \multicolumn{1}{c|}{89}  & 86  & 88        & 88       & 88     & 90        & 89       & 89      & 90       \\ \cline{3-21} 
                                                                           & \textbf{100}                   & 99       & 99     & 99    & 100      & 99      & 99     & 98       & 98      & 98    & \multicolumn{1}{c|}{99}   & \multicolumn{1}{c|}{100} & 99  & 99        & 99       & 99     & 99        & 99       & 99      & 99       \\ \hline
\end{tabular}
\end{adjustbox}
\label{tab:noise_tests}
\end{table*}

Gaussian Blur applied with different kernel sizes ($3\times3$, $9\times9$, $15\times15$) could destroy different main frequencies in the images. This filtering preserves low frequencies by almost totally deleting the high frequencies, as the kernel size increases. It is possible to see in Table~\ref{tab:noise_tests}, that the accuracy decreases at increasing of the kernel size. This phenomenon, is particularly visible for images generated by AttGAN, GDWCT and StarGAN which have the lowest resolution. 

Several degrees of rotation ($45, 90, 135, 180, 255$) were considered since they can modify the frequency components of the images. Rotations with angles of $90$, $180$, and $270$ do not alter the frequencies  because the [x,y] pixels are simply moved to the new [x',y'] coordinates without performing any interpolation function, obtaining  high values of detection accuracy. On the other hand, when considering different degrees of rotation, it is necessary to interpolate the neighboring pixels to get the missing ones. In this latter case, new information is added to the image that can affect the frequency information. In fact, considering rotations of 45, 135, 225 degree, the classification accuracy values decrease; except for the two StyleGANs for the same reason described for the Gaussian filter (i.e., high resolution).

The mirror attack reflects the image pixels along one axis (horizontal, vertical and both). This does not alter image frequencies, obtaining extremely high accuracy detection values.

The resizing attacks equal to $-$50\% of resolution causes a loss of information, hence, already small images tend to totally lose high-frequency components presenting a behavior similar to low-pass filtering; in this case accuracy values are inclined to be low. Vice versa, a resizing of +50\% doesn't destroy the main frequencies obtaining a high classification accuracy values. 

Finally, different JPEG compression quality factors were applied ($QF = 1, 50, 100$). As expected in Table~\ref{tab:noise_tests}, a compression with $QF = 100$ does not affect the results. 
The overall accuracy begins to be affected as the QF decreases, among other things, destroying the DCT coefficients. However, at $QF=50$ the mid-level frequencies are still preserved and the results maintain a high level of accuracy. This is extremely important given that this level of QF is employed by the most common social platforms such as Whatsapp or Facebook, thus demonstrating that the CTF approach is extremely efficient in real-world scenarios.

\subsection{Comparison and Generalization Tests}
\label{sec:comparison}

The CTF approach is extremely simple, fast, and demonstrates a high level of accuracy even in real-world scenarios. In order to better understand the effectiveness of the technique, a comparison with state-of-the-art methods was performed and reported in this section. The trained robust classifier was compared to the most recent work in the literature and in particular Zhang et al.~\cite{zhang2019detecting} (AutoGAN), Wang et al.~\cite{wang2019fakespotter} (FakeSpotter) and Guarnera et al.~\cite{guarnera2020fighting} (Expectation-Maximization) were considered for the use of a few GAN architectures in common with the analysis performed in this paper: StyleGAN, StyleGAN2, StarGAN. Table~\ref{tab:Compare1} shows that the CTF approach achieves the best results with an unbeatable accuracy of 99.9\%. 

\begin{table}[t!]
\centering
\caption{Comparison with state-of-the-art methods (\cite{zhang2019detecting, wang2019fakespotter,guarnera2020fighting}). Classification of Real images (CelebA and FFHQ) vs. Deepfake images. Accuracy values (\%) of each classification task are reported.}
\begin{tabular}{c|c|c|c|}
\cline{2-4}
\multicolumn{1}{l|}{}                      & \textbf{StarGAN} & \textbf{StyleGAN} & \textbf{StyleGAN2} \\ \hline
\multicolumn{1}{|c|}{\textbf{AutoGAN~\cite{zhang2019detecting}}}     & 65.6             & 79.5              & 72.5               \\ \hline
\multicolumn{1}{|c|}{\textbf{FakeSpotter~\cite{wang2019fakespotter}}} & 88               & 99.1              & 91.9               \\ \hline
\multicolumn{1}{|c|}{\textbf{EM~\cite{guarnera2020fighting}}}          & 90.55            & 99.48             & 99.64              \\ \hline
\multicolumn{1}{|c|}{\textbf{CTF (our)}}   &  99.9                & 100                  & 100                    \\ \hline
\end{tabular}

  \label{tab:Compare1}
\end{table}

Another comparison was made on the detection of StyleGAN and StarGAN with respect to \cite{guarnera2020fighting,wang2020cnn}. The obtained results are shown in the Table~\ref{tab:Compare3} in which the average classification values of each classification task are reported.

A specific discussion is needed for testing the FaceForensics++ dataset~\cite{rossler2019faceforensics++} which is a challenging dataset of fake videos of people speaking and acting in several contexts. The fake videos were created by means of four different techniques (Face2Face  \cite{thies2016face2face} among them) on videos taken from YouTube. 

By means of OpenCV's face detectors, cropped images of faces were taken from fake videos of FF++ (with samples from all four categories, at different compression levels) and a dataset of 3000 images with different resolutions ($min_{resolution} = 162 \times 162$ px, $max_{resolution} = 895 \times 895$ px). The CTF approach was employed to construct the $\beta$ feature vector computed on the DCT coefficients and the robust classifier (trained in the Section~\ref{sec:robustness}), was used for binary classification in order to perform this ``in the wild'' test.  
We emphasize that the latter datasets were only used in the testing phase with the robust classifier. 
Since the classifier detected FaceForensics++ images as well as StyleGAN images, we also tried to calculate the $GSF$ by comparing FaceForensics++ images with FFHQ obtaining a value of $61$ which is extremely close to the $GSF$ of StyleGANs. This leads to the explanation that the $GSF$s are also dependent not only {on} the generative process but also to the reenactment phase done on images. The reenactment is done analytically in Face2Face and trained in StyleGANs as a part of the model (similarly to Face2face but as a cost function).

\begin{table}[t!]
\centering
\caption{Comparison with state-of-the-art methods (\cite{wang2020cnn, guarnera2020fighting}). Classification of  Real images (CelebA and FFHQ) vs.  Deepfake images. The CTF approach was tested and compared  also considering the dataset of Deepfake video’s FaceForensics++ (FF++). Average Precision values (\%) of each classification task are reported.}
\begin{tabular}{c|c|c|c|}
\cline{2-4}
\multicolumn{1}{l|}{}                               & \textbf{StyleGAN} & \textbf{StarGAN} &  \textbf{FF++}\\ \hline
\multicolumn{1}{|c|}{\textbf{Wang~\cite{wang2020cnn}}}         & 96.3              & 100        &  98.2    \\ \hline
\multicolumn{1}{|c|}{\textbf{EM~\cite{guarnera2020fighting}}}                   & 99                & 93     &    98.8      \\ \hline
\multicolumn{1}{|c|}{\textbf{CTF (our)}}            &         99.9          &     99.9        &  99.9   \\ \hline
\end{tabular}
  \label{tab:Compare3}
\end{table}

The results obtained on FaceForensics++ are reported in Table \ref{tab:Compare3} showing how the CTF approach is an extremely simple and fast method capable of beating the state-of-the-art even on datasets on which it has not been trained and being able to catch not only convolutional artefacts but also those created by reenactment phase which is an important part for the most advanced Deepfake techniques.

\section{Conclusions}
\label{sec:conclusion}
In this paper, the CTF approach was presented as a detection method for Deepfake images. The approach is extremely fast, explainable, and does not need intense computational power for training. By exploiting and analyzing the overall statistics of the DCT coefficients it is possible to discriminate among all known GAN’s by means of the GAN Specific Frequency band ($GSF$). The $GSF$ has many interesting properties demonstrated through empirical and visual analysis; among others it is possible to give some explainability to the underlying generation process, especially for forensics purposes. In order to achieve higher accuracy values, all frequency bands must be taken into account and the CTF approach is finalized by means of a G-boost classifier which demonstrated to be robust to attacks and able to generalize even in a dataset of Deepfake videos (FaceForensics++) not used during training. Further investigation could be carried out on $GSF$ frequencies in order to detect not only GAN artefacts but also information coming from the reenactment phase. Finally, the CTF approach could give useful suggestions for the $GSF$ analysis (explainability, etc.) in new scenarios with more challenging modalities (attribute manipulation, expression swap, etc.) and media (audio,video){.}

\label{sect:bib}
\balance
\bibliographystyle{IEEEtran}
\bibliography{main}

\begin{thebibliography}{10}
\providecommand{\url}[1]{#1}
\csname url@samestyle\endcsname
\providecommand{\newblock}{\relax}
\providecommand{\bibinfo}[2]{#2}
\providecommand{\BIBentrySTDinterwordspacing}{\spaceskip=0pt\relax}
\providecommand{\BIBentryALTinterwordstretchfactor}{4}
\providecommand{\BIBentryALTinterwordspacing}{\spaceskip=\fontdimen2\font plus
\BIBentryALTinterwordstretchfactor\fontdimen3\font minus
  \fontdimen4\font\relax}
\providecommand{\BIBforeignlanguage}[2]{{%
\expandafter\ifx\csname l@#1\endcsname\relax
\typeout{** WARNING: IEEEtran.bst: No hyphenation pattern has been}%
\typeout{** loaded for the language `#1'. Using the pattern for}%
\typeout{** the default language instead.}%
\else
\language=\csname l@#1\endcsname
\fi
#2}}
\providecommand{\BIBdecl}{\relax}
\BIBdecl

\bibitem{goodfellow2014generative}
I.~Goodfellow, J.~Pouget-Abadie, M.~Mirza, B.~Xu, D.~Warde-Farley, S.~Ozair,
  A.~Courville, and Y.~Bengio, ``Generative adversarial nets,'' in
  \emph{Advances in Neural Information Processing Systems}, 2014, pp.
  2672--2680.

\bibitem{vaccari2020deepfakes}
C.~Vaccari and A.~Chadwick, ``Deepfakes and disinformation: exploring the
  impact of synthetic political video on deception, uncertainty, and trust in
  news,'' \emph{Social Media+ Society}, vol.~6, no.~1, p. 2056305120903408,
  2020.

\bibitem{guarnera2020deepfake}
L.~Guarnera, O.~Giudice, and S.~Battiato, ``Deepfake detection by analyzing
  convolutional traces,'' in \emph{Proceedings of the IEEE/CVF Conference on
  Computer Vision and Pattern Recognition Workshops}, 2020, pp. 666--667.

\bibitem{guarnera2020preliminary}
L.~Guarnera, O.~Giudice, C.~Nastasi, and S.~Battiato, ``Preliminary forensics
  analysis of deepfake images,'' in \emph{2020 AEIT International Annual
  Conference (AEIT)}, 2020, pp. 1--6.

\bibitem{zhang2019detecting}
X.~Zhang, S.~Karaman, and S.-F. Chang, ``Detecting and simulating artifacts in
  gan fake images,'' in \emph{2019 IEEE International Workshop on Information
  Forensics and Security (WIFS)}.\hskip 1em plus 0.5em minus 0.4em\relax IEEE,
  2019, pp. 1--6.

\bibitem{oliva2001modeling}
A.~Oliva and A.~Torralba, ``Modeling the shape of the scene: A holistic
  representation of the spatial envelope,'' \emph{International journal of
  Computer Vision}, vol.~42, no.~3, pp. 145--175, 2001.

\bibitem{xu2020learning}
K.~Xu, M.~Qin, F.~Sun, Y.~Wang, Y.-K. Chen, and F.~Ren, ``Learning in the
  frequency domain,'' in \emph{Proceedings of the IEEE/CVF Conference on
  Computer Vision and Pattern Recognition}, 2020, pp. 1740--1749.

\bibitem{xu2019training}
Z.-Q.~J. Xu, Y.~Zhang, and Y.~Xiao, ``Training behavior of deep neural network
  in frequency domain,'' in \emph{International Conference on Neural
  Information Processing}.\hskip 1em plus 0.5em minus 0.4em\relax Springer,
  2019, pp. 264--274.

\bibitem{NEURIPS2019_b05b57f6}
D.~Yin, R.~Gontijo~Lopes, J.~Shlens, E.~D. Cubuk, and J.~Gilmer, ``A fourier
  perspective on model robustness in computer vision,'' in \emph{Advances in
  Neural Information Processing Systems}, vol.~32, 2019, pp. 13\,276--13\,286.

\bibitem{rahaman2019spectral}
N.~Rahaman, A.~Baratin, D.~Arpit, F.~Draxler, M.~Lin, F.~Hamprecht, Y.~Bengio,
  and A.~Courville, ``On the spectral bias of neural networks,'' in
  \emph{International Conference on Machine Learning}.\hskip 1em plus 0.5em
  minus 0.4em\relax PMLR, 2019, pp. 5301--5310.

\bibitem{farinella2015representing}
G.~M. Farinella, D.~Rav{\`\i}, V.~Tomaselli, M.~Guarnera, and S.~Battiato,
  ``Representing scenes for real-time context classification on mobile
  devices,'' \emph{Pattern Recognition}, vol.~48, no.~4, pp. 1086--1100, 2015.

\bibitem{ravi2016semantic}
D.~Rav{\`\i}, M.~Bober, G.~M. Farinella, M.~Guarnera, and S.~Battiato,
  ``Semantic segmentation of images exploiting dct based features and random
  forest,'' \emph{Pattern Recognition}, vol.~52, pp. 260--273, 2016.

\bibitem{lam2000mathematical}
E.~Y. Lam and J.~W. Goodman, ``A mathematical analysis of the {DCT} coefficient
  distributions for images,'' \emph{IEEE Transactions on Image Processing},
  vol.~9, no.~10, pp. 1661--1666, 2000.

\bibitem{tolosana2020deepfakes}
R.~Tolosana, R.~Vera-Rodriguez, J.~Fierrez, A.~Morales, and J.~Ortega-Garcia,
  ``Deepfakes and beyond: A survey of face manipulation and fake detection,''
  \emph{arXiv preprint arXiv:2001.00179}, 2020.

\bibitem{verdoliva2020media}
L.~Verdoliva, ``Media forensics and deepfakes: an overview,'' \emph{IEEE
  Journal of Selected Topics in Signal Processing}, vol.~14, no.~5, pp.
  910--932, 2020.

\bibitem{choi2018stargan}
Y.~Choi, M.~Choi, M.~Kim, J.-W. Ha, S.~Kim, and J.~Choo, ``Stargan: Unified
  generative adversarial networks for multi-domain image-to-image
  translation,'' in \emph{Proceedings of the IEEE Conference on Computer Vision
  and Pattern Recognition}, 2018, pp. 8789--8797.

\bibitem{karras2019style}
T.~Karras, S.~Laine, and T.~Aila, ``A style-based generator architecture for
  generative adversarial networks,'' in \emph{Proceedings of the IEEE
  Conference on Computer Vision and Pattern Recognition}, 2019, pp. 4401--4410.

\bibitem{karras2020analyzing}
T.~Karras, S.~Laine, M.~Aittala, J.~Hellsten, J.~Lehtinen, and T.~Aila,
  ``Analyzing and improving the image quality of stylegan,'' in
  \emph{Proceedings of the IEEE/CVF Conference on Computer Vision and Pattern
  Recognition}, 2020, pp. 8110--8119.

\bibitem{he2019attgan}
Z.~He, W.~Zuo, M.~Kan, S.~Shan, and X.~Chen, ``Attgan: Facial attribute editing
  by only changing what you want,'' \emph{IEEE Transactions on Image
  Processing}, vol.~28, no.~11, pp. 5464--5478, 2019.

\bibitem{cho2019image}
W.~Cho, S.~Choi, D.~K. Park, I.~Shin, and J.~Choo, ``Image-to-image translation
  via group-wise deep whitening-and-coloring transformation,'' in
  \emph{Proceedings of the IEEE Conference on Computer Vision and Pattern
  Recognition}, 2019, pp. 10\,639--10\,647.

\bibitem{liu2015deep}
Z.~Liu, P.~Luo, X.~Wang, and X.~Tang, ``Deep learning face attributes in the
  wild,'' in \emph{Proceedings of the IEEE International Conference on Computer
  Vision}, 2015, pp. 3730--3738.

\bibitem{langner2010presentation}
O.~Langner, R.~Dotsch, G.~Bijlstra, D.~H. Wigboldus, S.~T. Hawk, and
  A.~Van~Knippenberg, ``Presentation and validation of the radboud faces
  database,'' \emph{Cognition and Emotion}, vol.~24, no.~8, pp. 1377--1388,
  2010.

\bibitem{zhu2017unpaired}
J.-Y. Zhu, T.~Park, P.~Isola, and A.~A. Efros, ``Unpaired image-to-image
  translation using cycle-consistent adversarial networks,'' in
  \emph{Proceedings of the IEEE International Conference on Computer Vision},
  2017, pp. 2223--2232.

\bibitem{lee2018diverse}
H.-Y. Lee, H.-Y. Tseng, J.-B. Huang, M.~Singh, and M.-H. Yang, ``Diverse
  image-to-image translation via disentangled representations,'' in
  \emph{Proceedings of the European Conference on Computer Vision (ECCV)},
  2018, pp. 35--51.

\bibitem{wilber2017bam}
M.~J. Wilber, C.~Fang, H.~Jin, A.~Hertzmann, J.~Collomosse, and S.~Belongie,
  ``Bam! the behance artistic media dataset for recognition beyond
  photography,'' in \emph{Proceedings of the IEEE International Conference on
  Computer Vision}, 2017, pp. 1202--1211.

\bibitem{durall2019unmasking}
R.~Durall, M.~Keuper, F.-J. Pfreundt, and J.~Keuper, ``Unmasking deepfakes with
  simple features,'' \emph{arXiv preprint arXiv:1911.00686}, 2019.

\bibitem{karras2017progressive}
T.~Karras, T.~Aila, S.~Laine, and J.~Lehtinen, ``Progressive growing of gans
  for improved quality, stability, and variation,'' \emph{arXiv preprint
  arXiv:1710.10196}, 2017.

\bibitem{wang2019fakespotter}
R.~Wang, L.~Ma, F.~Juefei-Xu, X.~Xie, J.~Wang, and Y.~Liu, ``Fakespotter: A
  simple baseline for spotting {AI}-synthesized fake faces,'' \emph{arXiv
  preprint arXiv:1909.06122}, 2019.

\bibitem{liu2019stgan}
M.~Liu, Y.~Ding, M.~Xia, X.~Liu, E.~Ding, W.~Zuo, and S.~Wen, ``Stgan: A
  unified selective transfer network for arbitrary image attribute editing,''
  in \emph{Proceedings of the IEEE Conference on Computer Vision and Pattern
  Recognition}, 2019, pp. 3673--3682.

\bibitem{rossler2019faceforensics++}
A.~Rossler, D.~Cozzolino, L.~Verdoliva, C.~Riess, J.~Thies, and M.~Nie{\ss}ner,
  ``Faceforensics++: Learning to detect manipulated facial images,'' in
  \emph{Proceedings of the IEEE International Conference on Computer Vision},
  2019, pp. 1--11.

\bibitem{li2020celeb}
Y.~Li, X.~Yang, P.~Sun, H.~Qi, and S.~Lyu, ``Celeb-df: A large-scale
  challenging dataset for deepfake forensics,'' in \emph{Proceedings of the
  IEEE/CVF Conference on Computer Vision and Pattern Recognition}, 2020, pp.
  3207--3216.

\bibitem{jain2020detecting}
A.~Jain, P.~Majumdar, R.~Singh, and M.~Vatsa, ``Detecting gans and retouching
  based digital alterations via dad-hcnn,'' in \emph{Proceedings of the
  IEEE/CVF Conference on Computer Vision and Pattern Recognition Workshops},
  2020, pp. 672--673.

\bibitem{ledig2017photo}
C.~Ledig, L.~Theis, F.~Husz{\'a}r, J.~Caballero, A.~Cunningham, A.~Acosta,
  A.~Aitken, A.~Tejani, J.~Totz, Z.~Wang \emph{et~al.}, ``Photo-realistic
  single image super-resolution using a generative adversarial network,'' in
  \emph{Proceedings of the IEEE Conference on Computer Vision and Pattern
  Recognition}, 2017, pp. 4681--4690.

\bibitem{radford2015unsupervised}
A.~Radford, L.~Metz, and S.~Chintala, ``Unsupervised representation learning
  with deep convolutional generative adversarial networks,'' \emph{arXiv
  preprint arXiv:1511.06434}, 2015.

\bibitem{pathak2016context}
D.~Pathak, P.~Krahenbuhl, J.~Donahue, T.~Darrell, and A.~A. Efros, ``Context
  encoders: Feature learning by inpainting,'' in \emph{Proceedings of the IEEE
  Conference on Computer Vision and Pattern Recognition}, 2016, pp. 2536--2544.

\bibitem{liu2020global}
Z.~Liu, X.~Qi, and P.~H. Torr, ``Global texture enhancement for fake face
  detection in the wild,'' in \emph{Proceedings of the IEEE/CVF Conference on
  Computer Vision and Pattern Recognition}, 2020, pp. 8060--8069.

\bibitem{hulzebosch2020detecting}
N.~Hulzebosch, S.~Ibrahimi, and M.~Worring, ``Detecting cnn-generated facial
  images in real-world scenarios,'' in \emph{Proceedings of the IEEE/CVF
  Conference on Computer Vision and Pattern Recognition Workshops}, 2020, pp.
  642--643.

\bibitem{guarnera2020fighting}
L.~Guarnera, O.~Giudice, and S.~Battiato, ``Fighting deepfake by exposing the
  convolutional traces on images,'' \emph{IEEE Access}, vol.~8, pp.
  165\,085--165\,098, 2020.

\bibitem{moon1996expectation}
T.~K. Moon, ``The expectation-maximization algorithm,'' \emph{IEEE Signal
  processing magazine}, vol.~13, no.~6, pp. 47--60, 1996.

\bibitem{jing2004face}
X.-Y. Jing and D.~Zhang, ``A face and palmprint recognition approach based on
  discriminant {DCT} feature extraction,'' \emph{IEEE Transactions on Systems,
  Man, and Cybernetics, Part B (Cybernetics)}, vol.~34, no.~6, pp. 2405--2415,
  2004.

\bibitem{thai2015camera}
T.~H. Thai, F.~Retraint, and R.~Cogranne, ``Camera model identification based
  on dct coefficient statistics,'' \emph{Digital Signal Processing}, vol.~40,
  pp. 88--100, 2015.

\bibitem{lam2004analysis}
E.~Y. Lam, ``Analysis of the dct coefficient distributions for document
  coding,'' \emph{IEEE Signal Processing Letters}, vol.~11, no.~2, pp. 97--100,
  2004.

\bibitem{dlib09}
D.~E. King, ``Dlib-ml: A machine learning toolkit,'' \emph{Journal of Machine
  Learning Research}, vol.~10, pp. 1755--1758, 2009.

\bibitem{wang2020cnn}
S.-Y. Wang, O.~Wang, R.~Zhang, A.~Owens, and A.~A. Efros, ``Cnn-generated
  images are surprisingly easy to spot... for now,'' in \emph{Proceedings of
  the IEEE Conference on Computer Vision and Pattern Recognition}, vol.~7,
  2020.

\bibitem{zhou2017scene}
B.~Zhou, H.~Zhao, X.~Puig, S.~Fidler, A.~Barriuso, and A.~Torralba, ``Scene
  parsing through ade20k dataset,'' in \emph{Proceedings of the IEEE Conference
  on Computer Vision and Pattern Recognition}, 2017, pp. 633--641.

\bibitem{bishop2006pattern}
C.~M. Bishop, \emph{Pattern Recognition and Machine Learning}.\hskip 1em plus
  0.5em minus 0.4em\relax Springer, 2006.

\bibitem{thies2016face2face}
J.~Thies, M.~Zollhofer, M.~Stamminger, C.~Theobalt, and M.~Nie{\ss}ner,
  ``Face2face: Real-time face capture and reenactment of rgb videos,'' in
  \emph{Proceedings of the IEEE Conference on Computer Vision and Pattern
  Recognition}, 2016, pp. 2387--2395.

\end{thebibliography}

\end{document}